\documentclass{article}

\usepackage{PRIMEarxiv}

\usepackage[utf8]{inputenc} 
\usepackage[T1]{fontenc}    
\usepackage{hyperref}       
\usepackage{url}            
\usepackage{booktabs}       
\usepackage{amsfonts}       
\usepackage{nicefrac}       
\usepackage{microtype}      
\usepackage{lipsum}
\usepackage{fancyhdr}       
\usepackage{graphicx}       
\graphicspath{{media/}}     

\usepackage{longtable}
\usepackage{float}
\usepackage{adjustbox}
\usepackage{multirow}
\usepackage{tabularx}
\usepackage{rotating} 
\usepackage{makecell}  
\usepackage{caption}
\usepackage{xcolor}

\usepackage{amsmath}
\usepackage{colortbl}

\usepackage{pifont}

\NewDocumentCommand\xtnote{sm}{\def\tss{\textsuperscript{#2}}\IfBooleanTF{#1}{\tss}{\rlap{\tss}}}

\newcolumntype{M}[1]{>{\centering\arraybackslash}m{#1}}

\title{Boosting Hyperspectral Image Classification with Gate-Shift-Fuse Mechanisms in a Novel CNN-Transformer Approach}

\author{
  M. F. Guerri\textsuperscript{1,2}, C. Distante\textsuperscript{1,2}, P. Spagnolo\textsuperscript{2}, F. Bougourzi\textsuperscript{2}, and A. Taleb-Ahmed\textsuperscript{3}\\
  \textsuperscript{1}Department of Innovation Engineering, University of Salento, 73100 Lecce, Italy\\
  \textsuperscript{2}National Research Council of Italy \\, Institute of Applied Sciences and Intelligent Systems, via Monteroni snc 73100 Lecce, Italy\\
  \textsuperscript{3}Institute d'Electronique de Microelectronique et  \\ de Nanotechnologie (IEMN), UMR 8520, Universite Polytechnique Hauts de France,  \\ Universite de Lille, CNRS, 59313, Valenciennes, France\\
  \texttt{\{mohamedfadhlallah.guerri, cosimo.distante\}@unisalento.it}, \\  \texttt{paolo.spagnolo@cnr.it}, \texttt{faresbougourzi@gmail.com}, \texttt{Abdelmalik.Taleb-Ahmed@uphf.fr}
}

\begin{document}

\maketitle

\begin{abstract}
During the process of classifying Hyperspectral Image (HSI), every pixel sample is categorized under a land-cover type. CNN-based techniques for HSI classification have notably advanced the field by their adept feature representation capabilities. However, acquiring deep features remains a challenge for these CNN-based methods. In contrast, transformer models are adept at extracting high-level semantic features, offering a complementary strength. This paper's main contribution is the introduction of an HSI classification model that includes two convolutional blocks, a Gate-Shift-Fuse (GSF) block and a transformer block. This model leverages the strengths of CNNs in local feature extraction and transformers in long-range context modelling. The GSF block is designed to strengthen the extraction of local and global spatial-spectral features. An effective attention mechanism module is also proposed to enhance the extraction of information from HSI cubes. The proposed method is evaluated on four well-known datasets (the Indian Pines, Pavia University, WHU-WHU-Hi-LongKou and WHU-Hi-HanChuan), demonstrating that the proposed framework achieves superior results compared to other models.

\end{abstract}


\section{Introduction}\label{Introduction}

Recent advancements in spectrum imaging technology have markedly elevated the precision of data collected through hyperspectral sensors. Hyperspectral imaging (HSI) is distinguished by its high-dimensional features and the richness and continuity of its spectral curves. HSI classification is used in various applications across fields such as agriculture \cite{gevaert2015generation}, biomedical imaging \cite{noor2016properties}, mineral exploration \cite{wang2012spectral}, food safety \cite{fong2020farm}, and military reconnaissance \cite{ardouin2007demonstration}.

Throughout the evolution of HSI classification, an extensive range of traditional methodologies has been proposed to refine classification accuracy. At their inception, HSI classification strategies were chiefly anchored in spectral linear transformations, notably principal component analysis (PCA) \cite{beirami2020band}, discriminant analysis \cite{imani2014principal}, \cite{li2011locality}, and independent component analysis (ICA) \cite{xia2016spectral}. These initial approaches tapped into the vast spectral data reservoir to categorize a variety of land covers effectively. In contrast, methods based on spectral classifiers, which are esteemed for their enhanced generalization abilities and nonlinear expression, have proven to be more effective in HSI classification. This genre includes sophisticated algorithms like k-nearest neighbour \cite{cariou2016new}, random forest \cite{xia2017random}, and support vector machine (SVM) \cite{chen2021feature}. Nonetheless, owing to the prevalent high intra-class spectral variability and minimal inter-class spectral differentiation, spectral-based techniques often falter in accurately distinguishing between disparate entities and are vulnerable to unanticipated noise.

Traditional machine learning techniques face challenges in feature extraction, modeling nonlinear relationships, and managing high-dimensional data in HSI \cite{li2019deep, guerri2024deep}. On the other hand, Deep Learning (DL) presents benefits like the automatic learning of features, powerful nonlinear modeling, efficient data representation, and data augmentation for better generalization \cite{bougourzi2023pdatt, guerri2024deep, bougourzi2023deep}. These advantages position DL as highly effective for the complex, high-dimensional, and nonlinear nature of HSI data, thereby significantly improving classification accuracy and robustness.

Due to their potent ability to extract both local spatial and spectral features, Convolutional Neural Networks (CNNs) have exhibited remarkable efficacy in HSI classification tasks. To address the challenges posed by high-dimensional HSI data, a Multiscale 3D deep convolutional neural network (M3D-DCNN) was introduced by \cite{he2017multi}, leveraging 3-D CNNs to mine spatial and spectral features within HSI data effectively. Additionally, the advent of more streamlined CNN models has been facilitated through the integration of group convolution techniques and attention mechanisms. Haut et al. \cite{haut2019visual} innovatively integrated CNNs and Residual Networks (ResNets) with visual attention to enhance feature identification, pinpointing the most significant parts of the data. This approach was proven by experimental evidence to endow deep attention models with a considerable competitive edge. Sun et al. \cite{sun2019spectral}, proved that the efficacy of CNN-based methods was compromised by interfering pixels, which diluted the discriminative capacity of spatial-spectral features, introduced a Spectral-Spatial Attention Network (SSAN). This network adeptly isolates discriminative spatial-spectral features within HSI's focused attention areas.

In this paper, we propose a CNN-Transformer approach for HSI data classification. The proposed approach exploits two types of convolutional operations (2D and 3D), a Gate-shift fuse (GSF) block and a Transformer block. The proposed approach has high efficiency for HSI data classification by employing the strengths of CNNs blocks for extracting local feature and the Transformer blocks for long-range context modelling. The main contributions of this paper can be summarized as follows:

\begin{itemize}
    \item We introduced the GSF designed specifically for HSI classification; GSF can strengthen CNNs and transformers in extracting local and global spatial-spectral features.
    \item We propose an effective attention mechanism module which enables better extraction of local and global information from HSI cubes. 
    \item We conduct a comparison analysis on four well-known datasets, where the proposed framework obtained superior results compared to other models.

\end{itemize}

The remainder of this article is organized as follows. The proposed model is introduced in Section \ref{methods}. Section \ref{experiments} shows the experimental datasets, the design of experimental parameters, and the comparison of classification accuracies. The discussion and some related conclusions are presented in Sections \ref{rworks}, \ref{disc}, \ref{conc}.

\section{Related Works}\label{rworks}

In \cite{sun2019spectral}, S. Hao et al. innovatively designed a lightweight spectral-spatial attention network (LSSAN) for HSI classification, achieving high accuracy with a substantial reduction in computational demand. Further, \cite{gao2020multiscale} introduced a multi-scale residual network (MSRN) that employs mixed depth-wise separable convolution for HSI classification, offering enhanced results with reduced computation. CNNs have significantly advanced HSI classification performance, but their fixed convolutional kernel sizes pose a challenge in capturing the long-range dependencies within the intricate HSI cube. Consequently, achieving superior classification performance often necessitates additional CNN layers, which runs counter to the objective of lightweight modeling. With the introduction of the vision transformer (ViT) model \cite{dosovitskiy2020image} and its applications in image processing, transformer-based networks have also found their way into HSI classification \cite{hong2021spectralformer}. The SpectralFormer (SF) network, detailed in \cite{hong2021spectralformer}, adeptly learns spectrally local sequence information through group-wise spectral embeddings. Additionally, \cite{mei2022hyperspectral} developed the group-aware hierarchical transformer (GAHT), which utilizes a grouped pixel embedding module to analyze spatial-spectral context in HSI data. There's also a growing trend among researchers to leverage CNNs for extracting shallow features before deploying transformers.

The study by Cui et al. \cite{cui2023madanet} presents MADANet, a lightweight network employing multi-scale feature aggregation and a dual attention mechanism. MADANet leverages depth-wise separable convolution for extracting multi-scale features, significantly enhancing local contextual information capture. Concurrently, its dual attention mechanism optimizes channel and spatial dimensions, providing a comprehensive understanding of global semantic information. MADANet's efficacy is demonstrated in scenarios with varying complexity, from agricultural to urban settings.

In \cite{zhang2023multi}, Lan Zhang et al. presented a transformer-based framework, Multi-Range Spectral-Spatial Transformer (MRSST), for HSI classification. It introduces a strategy for constructing composite token sequences, significantly enhancing the transformer's performance. The proposed MRSST architecture includes a convolutional feature pre-encoder with dual branches to capture localized features within each spectral band and a composite token generator that amalgamates spectral information and encoded features. This process results in token sequences enriched with multi-range information, enabling a more meticulous analysis of spectral band interdependencies. Experiments on three well-known datasets demonstrate that the multi-range composite token sequences and information exchange mechanisms notably improve transformer performance.

To conclude, utilizing transformer backbone networks in HSI classification is a viable approach, thanks to their enhanced global receptive field and superior ability to extract features from sequential data. Nonetheless, a notable downside is the substantial computational expense involved, as every position needs to be calculated at each process step.

\section{Methods}\label{methods}

\begin{figure*}[h!]
  \centering
  \includegraphics[width=\linewidth]{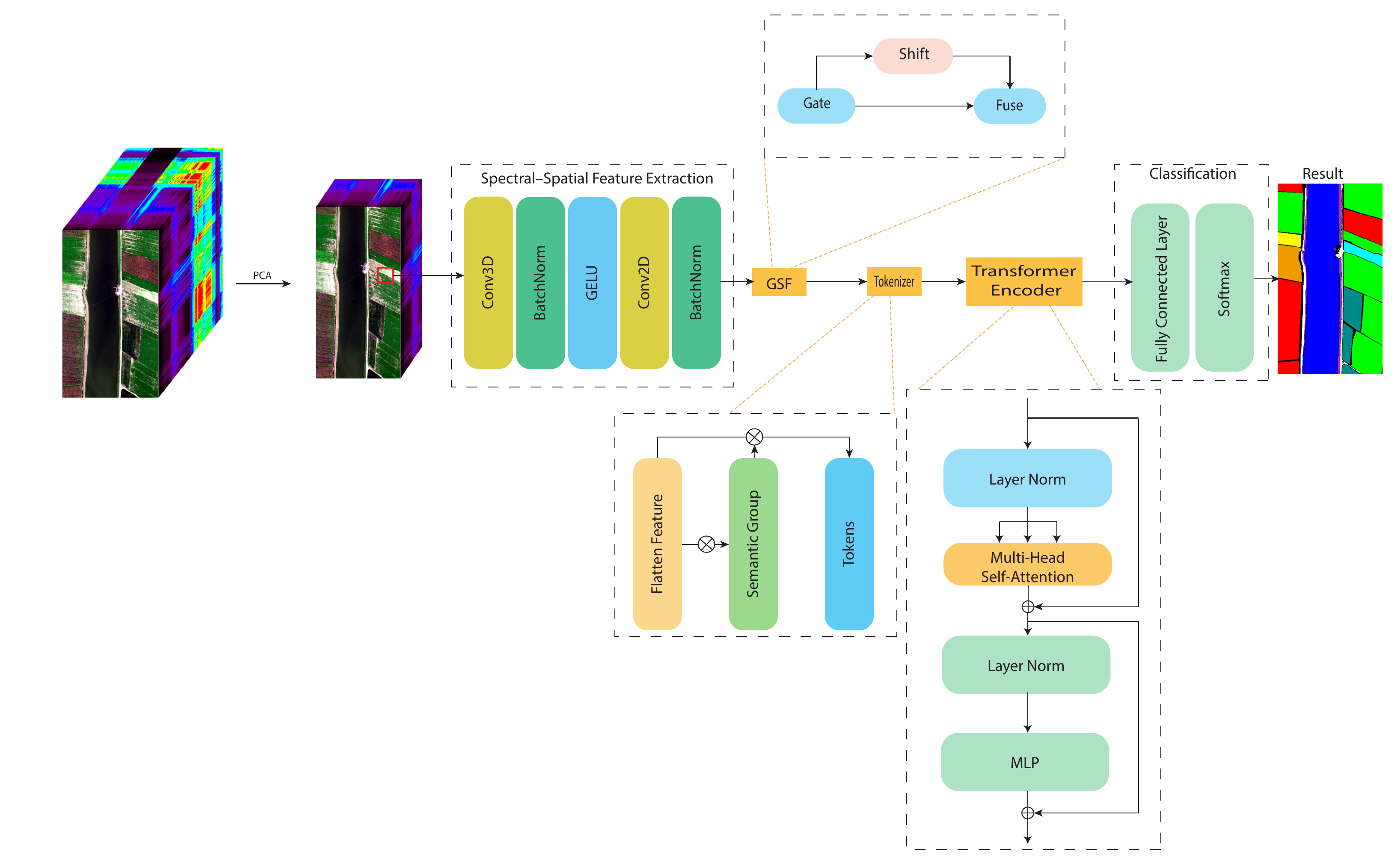}
  \caption{The proposed framework for the HSI classification, The process begins with PCA to reduce the spectral dimensionality of the HSI data. Then, the data undergo a spatial feature extraction phase using 3D and 2D convolution layers (Conv3D, Conv2D). The extracted features are then processed through the GSF block, enhancing the local and global feature representation. A tokenizer is used to convert the features into a sequence of tokens, which are fed into a transformer encoder used to capture long-range dependencies and high-level semantic features. Finally, the output passes through a linear layer and a softmax activation to classify the hyperspectral pixels, with the classification result presented as a color-coded map.}
  \label{fig:framework}
\end{figure*}

Fig \ref{fig:framework} shows the overall framework for the HSI classification based on the proposed model composed of four parts—the spectral–spatial feature extraction, GSF, the Gaussian weighted feature tokenizer, and the transformer-encoder (TE) module.

\subsection{Spectral–Spatial Feature Extraction}

The initial HSI dataset is represented as \(I \in \mathbb{R}^{m \times n \times l}\), indicating a spatial grid of \(m \times n\) with \(l\) spectral bands. In this configuration, each pixel in \(I\) encompasses \(l\) spectral values, constituting a one-hot category vector \(Y = (y_1, y_2, \ldots, y_C) \in \mathbb{R}^{1 \times 1 \times C}\) to represent the total \(C\) land-cover classes. The extensive spectral bands render the HSI data computationally intensive. To mitigate this, Principal Component Analysis (PCA) is employed to streamline both computation and spectral dimensions by reducing the spectral bands from \(l\) to \(b\), producing a transformed dataset \(I_{\mathrm{pca}} \in \mathbb{R}^{m \times n \times b}\), where \(b\) signifies the reduced number of spectral bands.

Subsequently, a 3-D patch extraction process is applied to \(I_{\mathrm{pca}}\), generating patches \(P \in \mathbb{R}^{s \times s \times b}\), where \(s \times s\) defines the patch dimensions. The position of a patch's center pixel is marked as \((x_i, x_j)\), constrained by \(0 \leq i < m\) and \(0 \leq j < n\), with the center pixel's label assigning the patch's true classification. To accommodate edge pixels and ensure full patch extraction, padding of width \((s-1)/2\) is applied. This methodology yields a total of \(m \times n\) 3-D patches, each spanning a width of \(x_i - (s-1)/2\) to \(x_i + (s-1)/2\) and a height of \(x_j - (s-1)/2\) to \(x_j + (s-1)/2\), inclusive of all \(b\) spectral bands.

Following the exclusion of patches with null labels, the residual patches are assorted into training and testing sets.

To extract detailed spectral-spatial characteristics from each patch, two convolution layers (3-D and 2-D) are utilized. Here, a 3-D convolution layer takes in training sample patches of dimension \(s \times s \times b\), with each layer's \(j\)-th feature cube's spatial position \((\alpha, \beta, \gamma)\) value being computed via a designated convolution operation.

\subsection{Gate-Shift-Fuse}

\begin{figure*}[h!]
  \centering
  \includegraphics[width=\linewidth]{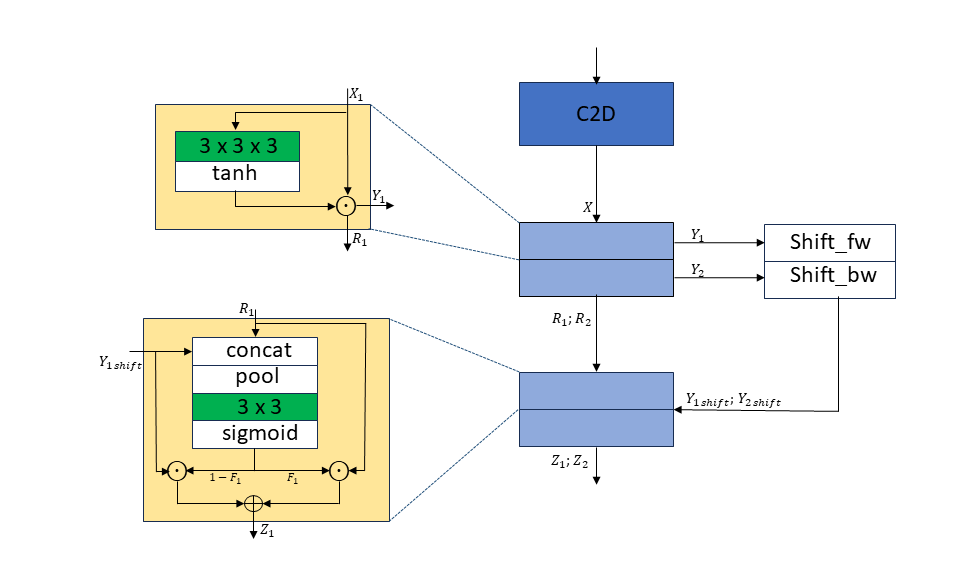}
  \caption{In the GSF framework, the integration of group gating, forward and backward spectral shift, and fusion mechanisms are employed. The gating mechanism is facilitated by a single 3D convolution kernel, finely tuned with tanh calibration. For the fusion process, a single 2D convolution kernel is used, refined with sigmoid calibration. As a result, the adoption of GSF introduces a negligible increase in parameters.}
  \label{fig:gsfarch}
\end{figure*}

Originating from \cite{sudhakaran2023gate}, the GSF framework features a spatial gating block and a subsequent fusion block, both of which are adaptable and designed to process dual splits of incoming feature sets. The gating block acts selectively, channeling features to emphasize spectral details. The fusion block, then, integrates these enriched spectral features with the spatial dimensions. Both components are dynamic, adjusting their operations based on the input data.

For the GSF operation, consider an input tensor \(X\) of dimensions \(C \times T \times W \times H\), where \(C\) is the channel count, and \(W\), \(H\), and \(T\) denote the spatial and spectral dimensions, respectively.

Following spatial convolution derived from a conventional 2D CNN model, \(X\) enters the gating mechanism. Here, it’s divided into two segments along the channel axis, creating two sets of gating planes, each \(1 \times T \times W \times H\), for the feature groups. These planes, acting as spatial weight maps, modulate the feature groups, resulting in gated features alongside their residuals. These are then spectrally shifted within their respective groups. These modified features and their residuals proceed to the fusion stage.

At the fusion stage, a channel-specific weight map of \(C \times T \times 1 \times 1\) for each group is generated. A weighted averaging process, leveraging these weight maps, fuses the shifted features and residuals. Through this process, GSF adeptly merges spatial and spectral data, guided by its trainable gating and fusion components.

\subsection{Gaussian-Weighted Feature Tokenizer}

While the GSF operation extracts features imbued with spectral and spatial attributes, these features alone fall short in comprehensively characterizing the attributes of terrestrial objects. Hence, to encapsulate and manipulate the advanced semantic constructs inherent to HSI feature categories, feature maps are reinterpreted as semantic tokens. This redefinition initiates with the input feature map, denoted as \(X \in \mathbb{R}^{u \times v \times z}\), where \(u\), \(v\), and \(z\) symbolize height, width, and channel count, respectively. Semantic tokens are articulated as \(T \in \mathbb{R}^{w \times z}\), with \(w\) signifying the token count.

The objective here is to transpose \(X\) into a semantic cohort, resulting in a semantic grouping represented by \(A \in \mathbb{R}^{u \times v \times w}\). Subsequently, \(A\) undergoes a transposition and is refined through the application of \(\text{softmax}(·)\) to accentuate segments of paramount semantic significance. The culmination of this process involves the multiplication of \(A\) with \(X\), thereby yielding \(T\) as the designated semantic tokens.

\subsection{Transformer Encoder Module}

\begin{figure*}[h!]
  \centering
  \includegraphics[width=\linewidth]{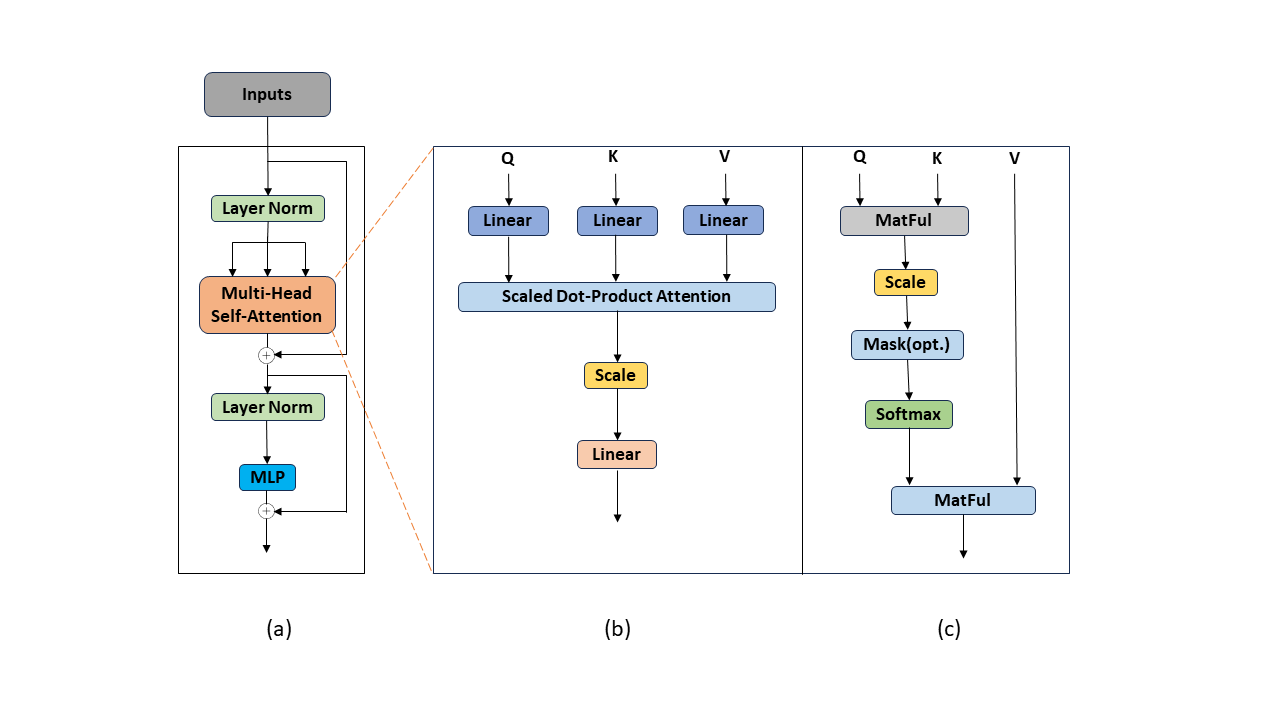}
  \caption{(a) The main Transformer Encoder framework. (b) Multi-Head Self-Attention (MSA). (c) Self-Attention (SA) module}
  \label{fig:temodule}
\end{figure*}

The TE module leverages semantic tokens generated earlier to discern relationships among high-level semantic features, as illustrated in Fig. \ref{fig:temodule}. This module is structured into three main components. The initial component employs position embedding to indicate the location of each semantic token, represented as $[T_1, T_2, \ldots, T_w]$. These tokens are merged with a learnable classification token $T_{\text{cls}}^0$, aimed at classification tasks. Following this, the positional information $PE_{\text{pos}}$ is encoded and merged with the token representations. The resulting sequence of embedded semantic tokens is:
\[
\text{Tin} = T_{\text{cls}}^0, T_1, T_2, \ldots, T_w + PE_{\text{pos}}. 
\]

The TE block, a pivotal component, is crafted to model intricate relationships among the semantic tokens. It includes a Multi-Head Self-Attention (MSA) block, an MLP layer, and two Normalization Layers (LN), enhanced by residual skip connections before the MSA and MLP layers.

The MSA block, the cornerstone of the transformer architecture, employs the Self-Attention (SA) mechanism to adeptly capture feature sequence correlations. It starts with defining three learnable matrices, $W_Q$, $W_K$, and $W_V$, for diverse interpretations. Mapping the tokens to these matrices forms 3-D matrices of queries $Q$, keys $K$, and values $V$. The attention score is derived from $Q$ and $K$, with the softmax function allocating the weights. SA is defined as:
\[
SA = \text{Attention}(Q, K, V) = \text{softmax}\left(\frac{QK^T}{\sqrt{d_K}}\right)V, 
\]
with $d_K$ being the key's dimension.

The MSA block maps $Q$, $K$, and $V$ using several weight matrix groups, performing the same operation to compute multi-head attention. The outcomes are concatenated:
\[
\text{MSA}(Q, K, V) = \text{Concat}(SA_1, SA_2, \ldots, SA_h)W, 
\]
with $h$ indicating the head count, and $W$ the parameter matrix, sized at $h \times d_K \times d_w$ ($d_w$ is the token count).

Subsequently, this matrix enters an MLP layer with two fully connected layers and a Gaussian error linear unit in-between. Following this, LN layers aid in training speed and mitigate gradient issues.

The TE module ensures the input ($T_{\text{in}}$) and output ($T_{\text{out}}$) sizes match. The classification token $T_{\text{out}}$ goes to the final classifier's linear layer, where a softmax function computes the input's category probability, with the highest value determining the category.

\subsection{Implementation}

This model distinguishes itself from traditional CNN frameworks by minimizing the layer count, introducing the GSF, a tokenizern and TE to interpret the semantic levels of image patches. We showcase this with the Pavia University dataset, sized \(610 \times 340 \times 103\).

Post-PCA reduction and patch extraction, patches are \(13 \times 13 \times 30\). The foremost 3-D convolution layer produces eight \(11 \times 11 \times 28\) feature cubes, using eight \(3 \times 3 \times 3\) cube kernels on each patch, harnessing the extensive spectral information present. These cubes are merged into one \(11 \times 11 \times 224\) cube. Then, 2-D convolution with \(64\) plane kernels of \(3 \times 3\) generates \(64\) \(9 \times 9\) feature maps, each flattened into a 1-D \(81\)-element vector, culminating in \(X \in \mathbb{R}^{81 \times 64}\).

Next, utilizing the Xavier normal distribution, we derive the initial weight matrix \(Wa \in \mathbb{R}^{64 \times 4}\) to normalize feature distribution. This matrix is applied to the feature vectors, producing the semantic array \(A \in \mathbb{R}^{81 \times 4}\). The transposition of \(A\) with \(X\) forms semantic tokens \(T \in \mathbb{R}^{4 \times 64}\). Integrating a zero vector into \(T\) as an adjustable token with position markers, we obtain \(T_{\text{in}} \in \mathbb{R}^{5 \times 64}\), which, after TE module processing, visualizes semantic features. The primary classification token output, \(T_{\text{out}}^0 \in \mathbb{R}^{1 \times 64}\), is used as the classifier's input to finalize the label.

\section{Experiment and Analysis} \label{experiments}
\subsection{Data Description}

Our method's performance was gauged through tests conducted on four Hyperspectral Imaging (HSI) datasets: Indian Pines, Pavia University, WHU-WHU-Hi-LongKou, and WHU-Hi-HanChuan.
Captured by the AVIRIS Sensor in 1992 over Indiana, USA, the Indian Pines dataset contains 224 spectral bands ranging from 0.4 to 2.5 um, covering 145×145 pixels at 20m resolution, featuring 16 land-cover classes. After eliminating 24 bands due to noise and water absorption, we selected 200 bands for experimentation, as illustrated in the ground-truth map (see Fig. [Reference]).

The Pavia University dataset, collected by the ROSIS Sensor in 2001 over Northern Italy, includes 115 spectral bands (0.43 to 0.86 um) across a 610×340 pixel area with a 1.3m resolution, categorizing nine land-cover types. Here, we discarded 12 noise bands, working with 103 bands.

Sourced from Hubei, China, using the Headwall NanoHyperspec sensor, the WHU-Hi datasets (WHU-Hi-LongKou and WHU-Hi-HanChuan) focus on agricultural lands. WHU-Hi-LongKou, with 9 classes and 270 spectral bands, covers 550×400 pixels at a 0.463m resolution. WHU-Hi-HanChuan offers 16 classes across 274 spectral bands, spanning 1217×303 pixels with a resolution of 0.109m.

\begin{figure}[h!]
    \centering
    \begin{minipage}[b]{0.28\textwidth} 
        \centering
        \includegraphics[width=\linewidth]{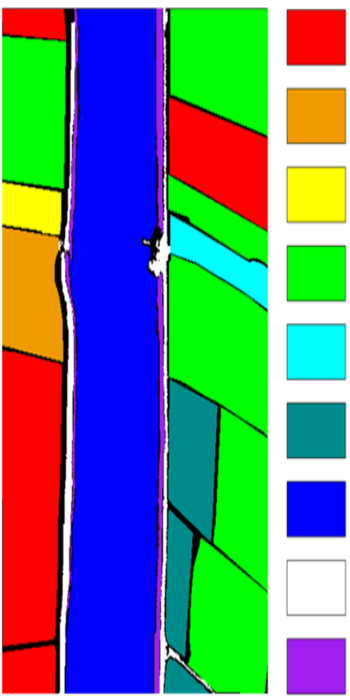}
        \captionof{figure}{Ground-truth image of WHU-Hi-LongKou dataset}
        \label{fig:lonngkou_gr}
    \end{minipage}%
    \hfill
    \begin{minipage}[b]{0.48\textwidth} 
        \centering
        \adjustbox{width=\linewidth}{
        \begin{tabular}{|c|p{4cm}|r|c|} 
          \hline
          \textbf{No.} & \textbf{Class name} & \textbf{Samples} & \textbf{Legend} \\
          \hline
          C1 & Corn & 34511 & \cellcolor[HTML]{ff0000} \\
          C2 & Cotton & 8374 & \cellcolor[HTML]{ef9b00} \\
          C3 & Sesame & 3031 & \cellcolor[HTML]{ffff00} \\
          C4 & Broad-leaf soybean & 63212 & \cellcolor[HTML]{00ff00} \\
          C5 & Narrow-leaf soybean & 4151 & \cellcolor[HTML]{00ffff} \\
          C6 & Rice & 11854 & \cellcolor[HTML]{008c8c} \\
          C7 & Water & 67056 & \cellcolor[HTML]{0000ff} \\
          C8 & Roads and houses & 7124 & \cellcolor[HTML]{ffffff} \\
          C9 & Mixed weed & 5229 & \cellcolor[HTML]{a020f0} \\
          \hline
        \end{tabular}
        }
        \captionof{table}{Groundtruth classes for the WHU-Hi-LongKou dataset}
        \label{tab:longkou_classes}
    \end{minipage}
\end{figure}

\begin{figure}[h!]
    \centering
    \begin{minipage}[b]{0.2\textwidth} 
        \centering
        \includegraphics[width=1\linewidth]{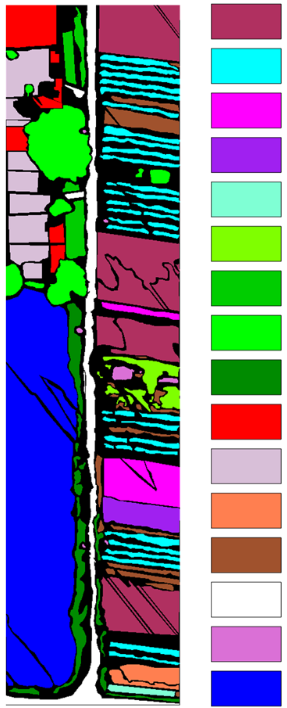}
        \caption{figure}{Ground-truth image of WHU-Hi-HuanChan dataset}
        \label{fig:huanchan_gr}
    \end{minipage}%
    \hfill
    \begin{minipage}[b]{0.48\textwidth} 
        \centering
        \adjustbox{max width=\textwidth}{
        \begin{tabular}{|c|p{4cm}|r|c|} 
          \hline
          \textbf{No.} & \textbf{Class name} & \textbf{Samples} & \textbf{Legend} \\
          \hline
          C1 & Strawberry & 44735 & \cellcolor[HTML]{b03060} \\
          C2 & Cowpea & 22753 & \cellcolor[HTML]{00ffff} \\
          C3 & Soybean & 10287 & \cellcolor[HTML]{ff00ff} \\
          C4 & Sorghum & 5353 & \cellcolor[HTML]{a020f0} \\
          C5 & Water spinach & 1200 & \cellcolor[HTML]{7fffd4} \\
          C6 & Watermelon & 4533 & \cellcolor[HTML]{7fff00} \\
          C7 & Greens & 5903 & \cellcolor[HTML]{00cd00} \\
          C8 & Trees & 17978 & \cellcolor[HTML]{00ff00} \\
          C9 & Grass & 9469 & \cellcolor[HTML]{008b00} \\
          C10 & Red roof & 10516 & \cellcolor[HTML]{ff0000} \\
          C11 & Gray roof & 16911 & \cellcolor[HTML]{d8bfd8} \\
          C12 & Plastic & 3679 & \cellcolor[HTML]{ff7f50} \\
          C13 & Bare soil & 9116 & \cellcolor[HTML]{a0522d} \\
          C14 & Road & 18560 & \cellcolor[HTML]{ffffff} \\
          C15 & Bright object & 1136 & \cellcolor[HTML]{da70d6} \\
          C16 & Water & 75401 & \cellcolor[HTML]{0000ff} \\
          \hline
        \end{tabular}
        }
        \captionof{table}{Groundtruth classes for the WHU-Hi-HuanChan dataset}
        \label{tab:huanchan_classes}
    \end{minipage}
\end{figure}

\subsection{Experimental Setting}

\subsubsection{Evaluation Indicators}

For a quantitative evaluation of our proposed method alongside others, we have employed four key quantitative indices: overall accuracy (OA), average accuracy (AA), kappa coefficient ($\kappa \times 100$), and classification accuracy across individual land-cover categories. Higher values in these metrics denote improved classification outcomes.

\subsubsection{Configuration}
The verification experiments for our proposed approach were conducted within the PyTorch framework, utilizing an NVIDIA DGX A100 Workstation equipped with Dual AMD Rome 7742 CPUs, 2TB RAM, and eight NVIDIA A100 80 GB GPUs. We opted for the Adam optimizer as the primary optimizer, setting the initial learning rate at 1e-3. For batch processing, minibatches were configured to contain 64 samples each. We applied 200 training epochs to each dataset.

\subsubsection{Classification Results}

To illustrate the proposed model's efficiency, a selection of representative methods was employed for comparison in experimental trials: SVM, EMAP \cite{dalla2010classification}, 1D-CNN \cite{hu2015deep}, 2D-CNN \cite{zhao2016spectral}, 3D-CNN \cite{chen2016deep}, SSRN \cite{zhong2017spectral}, Cubic-CNN \cite{wang2020novel}, HybridSN \cite{roy2019hybridsn} for Indian Pines and Pavia University Datasets, and 2DCNN \cite{zhao2016spectral}, 3DCNN \cite{chen2016deep}, FC3DCN \cite{ahmad2020fast}, HybridSN \cite{roy2019hybridsn}, S3EResBoF \cite{roy2020lightweight}, A2S2KRes \cite{roy2020attention}, FuSENet \cite{roy2020fusenet}, DLEM \cite{iyer2021deep} and our proposed method.

\begin{table*}[t]
  \centering
  \begin{adjustbox}{width=\textwidth}
    \begin{tabular}{cccccccccc}
      \hline
        NO. & SVM & EMAP \cite{dalla2010classification} & 1D-CNN \cite{hu2015deep} & 2D-CNN \cite{zhao2016spectral} & 3D-CNN \cite{chen2016deep} & SSRN \cite{zhong2017spectral} & Cubic-CNN \cite{wang2020novel} & DualMamba \cite{sheng2024dualmamba} & Ours \\
        \hline 
        1 & 65.63 & 62.50 & 43.75 & 48.78 & 41.46 & 83.15 & 87.86 & $\mathbf{96.83}$ & 95.12 \\
        2 & 63.44 & 81.57 & 77.93 & 78.13 & 90.51 & 95.31 & 96.35 & 99.07 & $\mathbf{99.67}$ \\
        3 & 60.25 & 83.19 & 56.72 & 83.51 & 79.36 & 94.23 & 93.65 & 99.00 & $\mathbf{99.18}$ \\
        4 & 41.11 & 85.89 & 45.18 & 47.42 & 46.01 & 90.68 & 82.54 & 98.29 & $\mathbf{98.55}$ \\
        5 & 87.05 & 78.61 & 87.57 & 75.12 & 95.17 & 97.79 & 96.69 & $\mathbf{97.91}$ & 96.32 \\
        6 & 97.21 & 79.08 & 98.63 & 92.99 & $\mathbf{99.70}$ & 98.67 & 96.69 & 99.54 & 99.54 \\
        7 & 89.47 & 52.63 & 65.11 & 60.00 & 88.00 & 97.92 & 90.16 & 99.85 & $\mathbf{100}$ \\
        8 & 96.66 & 91.19 & 97.36 & 98.37 & $\mathbf{100}$ & 99.26 & 98.46 & $\mathbf{100}$ & $\mathbf{100}$ \\
        9 & 32.26 & 50.12 & 37.14 & 66.67 & 48.89 & 89.49 & 89.93 & $\mathbf{100}$ & $\mathbf{100}$ \\
        10 & 73.84 & 81.32 & 66.03 & 87.77 & 86.06 & 97.48 & 93.94 & 98.29 & $\mathbf{99.16}$ \\
        11 & 84.36 & 86.91 & 82.49 & 89.09 & 97.51 & 98.16 & 97.45 & $\mathbf{99.58}$ & 98.69 \\
        12 & 42.89 & 78.43 & 73.49 & 63.67 & 74.91 & 93.07 & 93.18 & 99.01 & $\mathbf{99.13}$ \\
        13 & 98.58 & 96.35 & 99.30 & $\mathbf{100}$ & 99.46 & 98.59 & 99.12 & 99.73 & $\mathbf{100}$ \\
        14 & 94.02 & 93.91 & 93.78 & 95.33 & 99.74 & 99.72 & 99.39 & 99.72 & $\mathbf{99.91}$ \\
        15 & 42.65 & 77.36 & 55.39 & 66.76 & 84.10 & 93.31 & 84.26 & 92.35 & $\mathbf{98.84}$ \\
        16 & 92.19 & 84.38 & 81.54 & 91.57 & 93.98 & 93.79 & 89.69 & 97.98 & $\mathbf{99.08}$ \\
        \hline 
        OA(\%) & 76.39 & 83.69 & 79.37 & 84.47 & 91.03 & 94.78 & 94.90 & 99.23 & $\mathbf{99.47}$ \\
        AA(\%) & 72.18 & 76.53 & 70.87 & 77.83 & 82.18 & 94.67 & 93.85 & 99.11 & $\mathbf{99.57}$ \\
        $\kappa \times 100$ & 72.85 & 81.40 & 76.28 & 82.24 & 89.68 & 94.08 & 94.17 & 99.08 &$\mathbf{99.11}$ \\

        \hline
    \end{tabular}
  \end{adjustbox}
  \caption{Classification Results on Different Methods for the Indian Pines Dataset. Overall accuracy (OA), Average accuracy (AA) and Kappa Coefficient ($\kappa \times 100$).}
  \label{tab:ipresults}
\end{table*}

\begin{table*}[t]
  \centering
  \begin{adjustbox}{width=\textwidth}
    \begin{tabular}{cccccccccc}
      \hline
        NO. & SVM & EMAP \cite{dalla2010classification} & 1D-CNN \cite{hu2015deep} & 2D-CNN \cite{zhao2016spectral} & 3D-CNN \cite{chen2016deep} & SSRN \cite{zhong2017spectral} & Cubic-CNN \cite{wang2020novel} & HybridSN \cite{roy2019hybridsn} & Ours \\
        \hline 
         1 & 91.22 & 89.74 & 84.91 & 93.28 & 93.30 & 95.35 & 94.48 & 95.51 & $\mathbf{99.33}$ \\
         2 & 97.78 & 93.50 & 94.14 & 94.90 & 93.99 & 94.69 & 93.96 & 99.49 & $\mathbf{99.92}$ \\
         3 & 34.95 & 42.06 & 47.38 & 75.55 & 90.19 & 96.48 & 93.49 & 94.18 & $\mathbf{98.29}$ \\
         4 & 81.55 & 81.54 & 82.28 & 93.87 & 91.29 & 96.37 & 98.01 & $\mathbf{99.55}$ & 98.49 \\
         5 & 98.59 & 97.96 & 99.76 & 97.98 & 95.47 & 99.69 & $\mathbf{99.76}$ & 96.71 & 99.53 \\
         6 & 41.50 & 48.18 & 77.67 & 70.05 & 93.85 & 97.49 & 93.48 & 99.43 & $\mathbf{100.00}$ \\
         7 & 16.71 & 31.58 & 19.40 & 70.92 & 81.45 & 95.36 & 96.48 & $\mathbf{100.00}$ & 99.13 \\
         8 & 89.74 & 91.45 & 70.01 & 90.30 & 92.73 & 91.49 & 92.51 & 95.97 & $\mathbf{98.05}$ \\
         9 & $\mathbf{99.89}$ & 98.78 & 98.55 & 97.89 & 95.46 & 95.90 & 95.35 & 95.22 & 95.44 \\
        \hline 
         $\mathrm{OA}(\%)$ & 82.76 & 84.89 & 83.50 & 90.19 & 92.01 & 95.87 & 95.68 & 98.16 & $\mathbf{99.21}$ \\
         $\mathrm{AA}(\%)$ & 72.44 & 76.36 & 74.90 & 87.31 & 90.45 & 95.86 & 95.28 & 97.35 & $\mathbf{98.69}$ \\
         $\kappa \times 100$ & 76.28 & 80.41 & 77.90 & 87.52 & 90.87 & 95.78 & 95.55 & 97.57 & $\mathbf{99.15}$ \\
        \hline
    \end{tabular}
  \end{adjustbox}
  \caption{Classification Results on Different Methods for the Pavia University Dataset}
  \label{tab:puresults}
\end{table*}

\begin{table*}[h!]
  \centering
  \begin{adjustbox}{width=\textwidth}
    \begin{tabular}{cccccccccc}
      \hline
        NO. & 2DCNN \cite{zhao2016spectral} & 3DCNN \cite{chen2016deep} & FC3DCN \cite{ahmad2020fast} & HybridSN \cite{roy2019hybridsn} & S3EResBoF \cite{roy2020lightweight} & A2S2KRes \cite{roy2020attention} & FuSENet \cite{roy2020fusenet} & DLEM \cite{iyer2021deep} & Ours \\
        \hline 
         1 &  99.95 & 99.89 & 99.90 & 99.99 & 99.97 & 99.31 & 99.83 & $\mathbf{99.99}$ & 99.96 \\
         2 & 98.89 & 99.20 & 99.26 & 99.87 & 99.01 & 99.30 & 99.42 & 99.89 & $\mathbf{99.90}$ \\
         3 & 97.48 & 99.83 & 99.76 & 98.98 & 100 & 99.13 & 99.82 & 99.42 & $\mathbf{100}$ \\
         4 & 99.67 & 99.71 & 99.62 & 99.46 & 99.88 & 99.81 & 99.31 & 99.85 & $\mathbf{99.94}$ \\
         5 & 97.07 & 98.46 & 97.76 & 97.59 & 99.46 & $\mathbf{99.84}$ & 99.54 & 98.34 & 98.86 \\
         6 & 99.57 & 99.86 & 99.92 & 99.88 & 99.99 & 99.63 & 99.97 & 99.87 & $\mathbf{100}$ \\
         7 & 99.98 & 99.97 & 99.98 & 99.96 & 99.88 & 99.98 & $\mathbf{99.99}$ & 99.98 & 99.97 \\
         8 & 95.45 & 98.16 & 98.41 & 98.16 & 92.85 & 98.26 & 95.79 & 98.36 & $\mathbf{99.02}$ \\
         9 & 94.75 & 94.08 & 97.06 & 98.13 & $\mathbf{99.55}$ & 97.64 & 98.91 & 97.08 & 98.05 \\
        \hline 
         $\mathrm{OA}(\%)$ & 99.42 & 99.59 & 99.64 & 99.63 & 99.59 & 99.63 & 99.54 & 99.76 & $\mathbf{99.86}$ \\
         $\mathrm{AA}(\%)$ & 98.09 & 98.79 & 99.07 & 99.11 & 98.96 & 99.21 & 99.18 & 99.20 & $\mathbf{99.52}$ \\
         $\kappa \times 100$ & 99.24 & 99.46 & 99.64 & 99.52 & 99.46 & 99.51 & 99.40 & 99.68 & $\mathbf{99.81}$ \\
        \hline
    \end{tabular}
  \end{adjustbox}
  \caption{Classification Results on Different Methods for the WHU-HI LongKou}
  \label{tab:lkresults}
\end{table*}

\begin{table*}[t]
  \centering
  \begin{adjustbox}{width=\textwidth}
    \begin{tabular}{cccccccccc}
      \hline
        NO. & 2DCNN \cite{zhao2016spectral} & 3DCNN \cite{chen2016deep} & FC3DCN \cite{ahmad2020fast} & HybridSN \cite{roy2019hybridsn} & S3EResBoF \cite{roy2020lightweight} & A2S2KRes \cite{roy2020attention} & FuSENet \cite{roy2020fusenet} & DLEM \cite{iyer2021deep} & Ours \\
        \hline 
         1 &  96.61 & 96.80 & 98.46 & 97.19 & 91.61 & 95.26 & 97.49 & 98.41 & $\mathbf{98.98}$ \\
         2 & 93.65 & 93.14 & 95.73 & 94.38 & 95.07 & 97.44 & 97.16 & 96.69 & $\mathbf{99.62}$ \\
         3 & 86.85 & 91.03 & 89.49 & 89.96 & 92.12 & 94.71 & $\mathbf{99.09}$ & 94.38 & 98.82 \\
         4 & 98.21 & 97.82 & 98.46 & 98.28 & 98.14 & 97.67 & 99.61 & 98.77 & $\mathbf{99.94}$ \\
         5 & 69.59 & 67.18 & 74.74 & 61.43 & 96.94 & 89.86 & 87.30 & 90.55 & $\mathbf{99.91}$ \\
         6 & 68.41 & 67.36 & 69.82 & 70.00 & 79.46 & 78.51 & 57.36 & 75.96 & $\mathbf{95.18}$ \\
         7 & 88.02 & 90.24 & 87.53 & 87.63 & 93.61 & 91.80 & 93.58 & 93.24 & $\mathbf{99.43}$ \\
         8 & 87.54 & 90.41 & 94.72 & 91.68 & 94.11 & 92.96 & 92.87 & 95.30 & $\mathbf{98.55}$ \\
         9 & 89.13 & 89.01 & 90.58 & 92.75 & 92.31 & 92.99 & 91.03 & 94.28 & $\mathbf{98.01}$ \\
         10 & 99.01 & 97.80 & 98.13 & 98.99 & 98.33 & 98.84 & 96.78 & 98.67 & $\mathbf{99.35}$ \\
         11 & 96.74 & 96.96 & 98.85 & 98.25 & 92.37 & 96.42 & 94.90 & 97.51 & $\mathbf{98.83}$ \\
         12 & 66.74 & 87.31 & 80.44 & 69.36 & 84.98 & 84.95 & 57.56 & 82.18 & $\mathbf{97.85}$ \\
         13 & 77.10 & 75.30 & 81.08 & 75.67 & 76.54 & 86.51 & 67.25 & 84.27 & $\mathbf{97.46}$ \\
         14 & 90.05 & 91.82 & 87.45 & 91.65 & 91.31 & 98.42 & 98.74 & 94.68 & $\mathbf{99.61}$ \\
         15 & 57.89 & 73.23 & 84.94 & 72.41 & 96.10 & 91.42 & 97.43 & 84.93 & $\mathbf{99.99}$ \\
         16 & 99.33 & 99.73 & 99.71 & 99.38 & 99.95 & 99.78 & 99.65 & 99.76 & $\mathbf{99.84}$ \\
        \hline 
         $\mathrm{OA}(\%)$ & 93.40 & 94.28 & 95.09 & 94.41 & 94.27 & 96.87 & 94.05 & 96.47 & $\mathbf{99.23}$ \\
         $\mathrm{AA}(\%)$ & 85.30 & 87.82 & 89.38 & 86.94 & 92.06 & 92.97 & 89.24 & 92.47 & $\mathbf{98.47}$ \\
         $\kappa \times 100$ & 93.30 & 94.25 & 93.45 & 93.28 & 96.34 & 93.05 & 95.87 & 95.09 & $\mathbf{99.03}$ \\
        \hline
    \end{tabular}
  \end{adjustbox}
  \caption{Classification Results on Different Methods for the WHU-HI HuanChan}
  \label{tab:hcresults}
\end{table*}

The performance of our proposed method and comparative methods across the Indian Pines, Pavia University, and WHU-HI datasets is detailed in Tables \ref{tab:ipresults}, \ref{tab:puresults}, \ref{tab:lkresults}, and \ref{tab:hcresults}, assessing OA, AA, $\kappa$, and individual class accuracies. The optimal results are emphasized in bold, underscoring our method's superior performance in achieving the highest OA, AA, $\kappa$, and class accuracy values. Particularly, Table \ref{tab:ipresults} illustrates the challenges of unbalanced classification in specific categories like ``alfalfa,'' ``grass-pasture-mowed,'' and ``oats'' within SVM, EMAP, and 1-D-CNN methods due to limited sample sizes and the consequent sample imbalance from percentage sampling. Our method's uniform class accuracy highlights its efficacy in classifying small samples based on regional semantic concepts.

In SVM classification, the RBF kernel was used, determining the best hyperparameters $\lambda$ and $\lambda$ via cross-validation. The EMAP method set the number of principal components to three and selected attribute profiles based on area, moment of inertia, and standard deviation, with predetermined scales for each. The 1-D-CNN consisted of five layers, including 20 1-D convolutional filters in the convolution layer. The 2-D-CNN architecture featured three convolutional blocks and two linear layers.

Within the 3-D-CNN model, the network is comprised of three 3-D convolutional blocks and two linear layers, mirroring the structure of the 2-D-CNN. Each block within the 3-D-CNN setup includes a 3-D convolutional layer, a batch normalization (BN) layer, and a ReLU activation function, housing eight, 16, and 32 3-D filters sized \(3 \times 3\), respectively.

Specific configurations for SSRN, Cubic-CNN, and HybridSN are detailed in their individual references. For our SSFTT approach, we designated 30 principal components. The network’s 3-D convolution layer features eight kernels of \(3 \times 3 \times 3\) dimensions, and the 2-D convolution layer includes 64 kernels of \(3 \times 3\). Patch dimensions were set to \(13 \times 13\), with the sample token count and the number of TE heads both fixed at four. A random selection process was employed to compile training and test sets across all methods, ensuring an equitable comparison.

\subsection{Ablation Experiments}

\begin{table}[h!]
    \begin{adjustbox}{width=\columnwidth,center}
        \begin{tabular}{c|c|c|c|c|c|c|c}
            \hline
            \multirow{2}{*}{Cases} & \multicolumn{4}{|c|}{Component} & \multicolumn{3}{c}{Results} \\
            \cline{2-8}
            & 2D Conv & 3D Conv & GSF & TE & OA (\%) & AA (\%) & $\kappa \times 100$ \\
            \hline    
            1 & $\times$ & $\sqrt{}$ & $\sqrt{}$ & $\sqrt{}$ & 94.03 & 90.60 & 93.33 \\
            2 & $\sqrt{}$ & $\times$ & $\sqrt{}$ & $\sqrt{}$ & 85.54 & 69.83 & 80.58 \\
            3 & $\times$ & $\times$ & $\times$ & $\sqrt{}$ & 90.98 & 85.55 & 88.06 \\
            4 & $\sqrt{}$ & $\sqrt{}$ & $\times$ & $\sqrt{}$ & 97.51 & 94.55 & 95.51 \\
            5 & $\sqrt{}$ & $\sqrt{}$ & $\times$ & $\times$ & 93.52 & 92.19 & 91.33 \\
            \hline
            6 & $\sqrt{}$ & $\sqrt{}$ & $\sqrt{}$ & $\sqrt{}$ & $\mathbf{99.23}$ & $\mathbf{98.47}$ & $\mathbf{99.03}$ \\
            \hline
        \end{tabular}
    \end{adjustbox}
    \caption{Ablation Analysis of the Proposed Model on WHU-HI HuanChan dataset}
    \label{Tab:ablationtab}
    \end{table}

In order to thoroughly showcase the effectiveness of our method, we undertook ablation studies on the WHU-HI HuanChan dataset, exploring five different configurations of component combinations. This analysis aimed to discern the impact of various components on the model's overall classification accuracy, with all results detailed in Table \ref{Tab:ablationtab}. The analysis deconstructed the model into five principal components: the 3-D convolution layer, 2-D convolution layer, GSF, tokenizer, and TE. The absence of the 2-D convolution layer led to the least effective classification performance. Removing the 3-D convolution layer slightly improved outcomes compared to the absence of the 2-D layer. The model achieved 90.98\% accuracy in the third configuration. In the fourth, replacing GSF with PE and keeping both convolution layers intact resulted in a commendable 97.51\% accuracy, just shy of our method's outcome, highlighting the GSF's role in classification enhancement. The fifth configuration, relying solely on the convolution layers for spectral–spatial feature analysis, achieved 93.52\% accuracy, demonstrating the significant contribution of the TE module towards performance improvement. These ablation study outcomes further validate our model's robustness.

\color{black}
\subsection{Time Cost Comparison}
A comparison of training and testing time for 2-D-CNN, 3-D-CNN, SSRN, Cubic-CNN, HybridSN, and our method is shown in Table \ref{Tab:timecomp}.
\begin{table}[h!]
    \begin{adjustbox}{width=\columnwidth,center}
        \begin{tabular}{c|c|c|c|c|c|c}
        \hline \multirow{2}{*}{ Methods } & \multicolumn{2}{|c|}{ Indian Pines } & \multicolumn{2}{c|}{ Pavia University } & \multicolumn{2}{c}{WHU-HI HuanChan} \\
        \cline { 2 - 7 } & Train(m) & Test(s) & Train(m) & Test(s) & Train(m) & Test(s) \\
        \hline 2D-CNN \cite{zhao2016spectral} & $\mathbf{1.45}$ & $\mathbf{3.32}$ & $\mathbf{1.91}$ & $\mathbf{4.82}$ & $\mathbf{2 .19}$ & $\mathbf{4.34}$ \\
        3D-CNN \cite{chen2016deep} & 7.26 & 6.90 & 11.48 & 14.62 & 8.29 & 9.10 \\
        SSRN \cite{zhong2017spectral} & 11.74 & 8.83 & 14.46 & 24.43 & 8.93 & 11.75 \\
        Cubic-CNN \cite{wang2020novel} & 10.54 & 9.60 & 13.58 & 29.91 & 9.73 & 15.18 \\
        HybridSN \cite{roy2019hybridsn} & 6.61 & 7.93 & 9.32 & 28.40 & 6.73 & 8.49 \\
        Ours & $\underline{2.29}$ & $\underline{5.29}$ & $\underline{2.17}$ & $\underline{9.85}$ & $\underline{2.62}$ & $\underline{4.49}$ \\
        \hline
        \end{tabular}
    \end{adjustbox}
    \caption{Training time in minutes and test time is seconds between our method and other methods on three datasets}
\label{Tab:timecomp}
\end{table}

Our approach outpaces the competition in terms of speed, barring the 2-D-CNN, thus cutting down on computation time and boosting classification efficiency. Due to their more complex and deeper network structures, SSRN and Cubic-CNN experience longer durations in both training and testing phases, consuming extensive computational cycles for each iteration. Our method's marginally longer computation time than that of the 2-D-CNN can be credited to the employment of a 3-D convolution layer for extracting spectral features. Additionally, the time invested in residual calculations within the TE structure is noteworthy.

\color{black}
\section{Discussion} \label{disc}
In this study, we embarked on a journey to enhance HSI classification by ingeniously intertwining the strengths of CNNs and transformers within a unified architecture. Our proposed model, fortified with two convolutional blocks, a Gate-Shift-Fuse (GSF) block and a transformer block, has unveiled a new horizon in HSI classification, demonstrating remarkable prowess in harnessing local and global spatial-spectral features.

A pivotal cornerstone of our approach is introducing the GSF block, a novel mechanism meticulously designed to bolster the extraction of intricate spatial-spectral features. This innovation has proven instrumental in navigating the complexities and nuances inherent in HSI data, facilitating a more nuanced and robust feature extraction process. Coupled with an effective attention mechanism module, our model has exhibited a remarkable aptitude for discerning and leveraging critical local information within HSI cubes. Our model has shown a particular resilience in grappling with the challenges posed by unbalanced classification scenarios, showcasing its ability to maintain performance integrity even in the face of categories marked by smaller sample sizes.

In dissecting the performance landscapes, it becomes evident that the synergistic interplay between CNNs and transformers within our model is a powerful catalyst, driving enhanced classification outcomes. The CNN components, renowned for their local feature extraction prowess, seamlessly intertwine with the transformers' capacity for long-range context modelling, culminating in a harmonized architecture that resonates with the multifaceted nature of HSI data.

\section{Conclusion} \label{conc}

This paper presents a novel method designed to boost HSI classification performance by organically fusing a backbone CNN with a transformer architecture. The convolution layers, enhanced by the GSF block, excel at extracting low-level spectral–spatial features. Subsequently, these features are transformed into semantic tokens. The method employs the TE structure to process the high-level semantic attributes of the tokens, facilitating a comprehensive analysis of land-cover traits. Experimentally, this approach has been proven to elevate classification efficacy significantly. Furthermore, it convincingly illustrates the feasibility of expanding the scope of HSI classification into the realm of local high-level semantic classification.

\section*{Acknowledgement}
The authors thank Mr. Arturo Argentieri from CNR-ISASI Italy for his technical contribution to the multi-GPU computing facilities. This research was funded in part by Future Artificial Intelligence Research—FAIR CUP B53C220036 30006 grant number PE0000013, and in part by the Ministry of Enterprises and Made in Italy with the grant ENDOR "ENabling technologies for Defence and mOnitoring of the foRests" - PON 2014-2020 FESR - CUP B82C21001750005. 


\end{document}